# Cognitive Architecture for Decision-Making Based on Brain Principles Programming

Anton Kolonin[a], Andrey Kurpatov[b], Artem Molchanov[b], Gennadiy Averyanov[b]

*[a]Novosibirsk State University, Pirogova 2, Novosibirsk, 630090, Russia*
*[b]Sberbank of Russia, Neuroscience Lab, Vavilova 19, Moscow, 117312, Russia*

**Abstract**

We describe a cognitive architecture intended to solve a wide range of problems based on the five identified principles of brain activity, with their implementation in three subsystems: logical-probabilistic inference, probabilistic formal concepts, and functional systems theory. Building an architecture involves the implementation of a task-driven approach that allows defining the target functions of applied applications as tasks formulated in terms of the operating environment corresponding to the task, expressed in the applied ontology. We provide a basic ontology for a number of practical applications as well as for the subject domain ontologies based upon it, describe the proposed architecture, and give possible examples of the execution of these applications in this architecture.

*Keywords:* brain principles programming; cognitive architecture; formal concept analysis; functional system theory; probabilistic logic; subject domain ontology; task-driven approach;

## 1. The concept of the cognitive kernel – five principles, one approach, three components

We propose to build a universal decision-making system based on the 5 principles of brain activity according to "Brain Principles Programming" concept (BPP) [1,2] within the framework of a task-driven approach (TDA) [3,4,5]. The principles are implemented in a cognitive architecture defined for an arbitrarily given operating environment [6], operating on the basis of logical-probabilistic inference (LPI) [7], probabilistic formal concepts (PFT) [8] and the theory of functional systems (TFS) [3,4,5].

### 1.1. Five principles of brain activity

The five principles of brain activity formulated in [1,2] suggest the simultaneous involvement of all of them in most cognitive processes. That is, when making a decision in any new situation, the following is happening: 1) the principle of "complexity generation" leads to the generation of a number of hypotheses about the extended context of the situation for a limited number of stimuli or perceptions in various modalities; 2) the principle of "relationship" makes it possible to correlate different stimuli and hypotheses with each other, magnifying some and dismissing others; 3) the principle of "approximation to the essence" makes it possible to connect partial manifestations mediated by stimuli to concepts and phenomena known from accumulated past experiences; 4) the principle of "locality-distribution" ensures the "assembly" of full-fledged integrated contexts based on a set of particular stimuli and concepts of one or different modalities; 5) the principle of "heaviness" determines the dominant context corresponding to the most likely hypothesis – an example of this happening is shown in Fig.1.

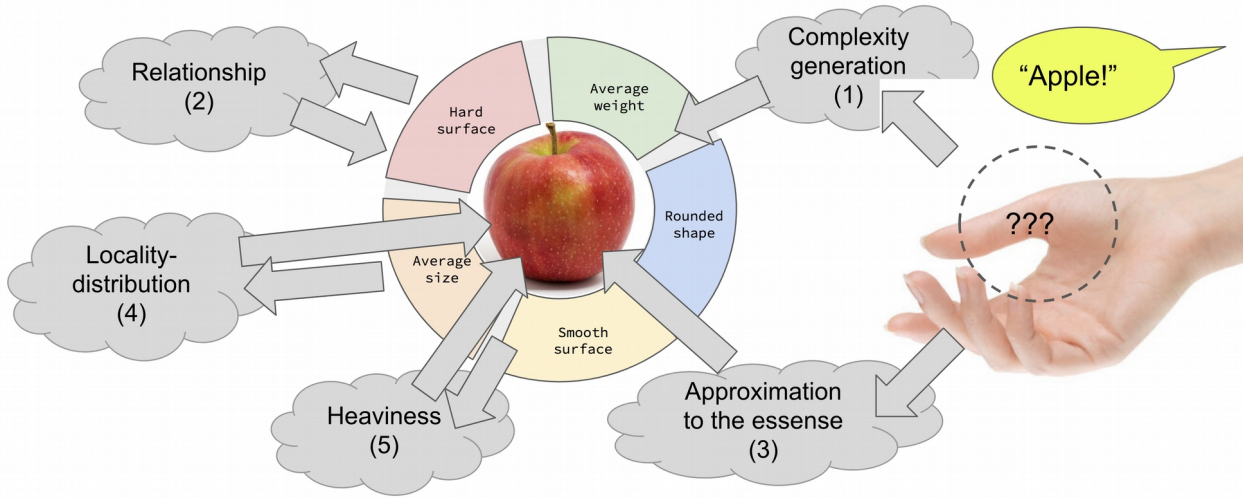

Fig. 1. Application of the five principles of brain activity to solving the cognitive problem of recognizing an invisible object by touch.

The sensation of a round object in the hand of the subject, with their eyes being closed, allows for the generation of a number of hypotheses. The unknown object, due to the principle of "complexity generation", is endowed with various properties, including those that are not yet confirmed, that are associated with a series of hypotheses (an orange orange, a white billiard ball, a red apple, a green apple). The "relationship" principle allows us to correlate these hypotheses with those properties that are actually perceived, allowing the subject to conclude that the softness of a hypothetical orange does not correspond to the hardness of the perceived object, just as the heaviness of a hypothetical billiard ball does not correspond with the perceived lightness. The smoothness and lightness of an apple, however, coincides with the actual perceived sensations. The principle of "locality-distribution" allows the subject to create a single image from multiple modalities: the feeling of a smooth and hard surface comes from the fingers and palms, the light weight – from the forearm and elbow, the apple smell – from the olfactory organs, information about being in a grocery store (and not in a billiard club) – from short-term memory. Due to the "heaviness" principle, the "apple" hypothesis becomes dominant, suppressing the "orange" and "billiard ball" hypotheses. Finally, the principle of "approximation to essence" crystallizes the image of a concretely tangible apple, in all its manifestations, to an abstract stereotyped apple in the subject's thoughts, allowing the subject to confidently give the correct answer "this is an apple!".

*1.2. Task-driven approach*

The task-driven approach (TDA) to the problem of artificial intelligence, formulated in [3,4,5] on the basis of the Theory of Functional Systems (TFS) by P.K. Anokhin, suggests the possibility of solving a wide class of cognitive tasks. A necessary condition for this the formalization of these tasks in the appropriate operational space in terms of the ontology of a particular subject area. The specified formalization should provide the possibility of describing in these terms both the problem to be solved and the image of an achievable result, taking into account the criteria for achieving the result. The latter can be done by introducing the function of "success" (of problem solving) or "utility" (of the subject solving the given task or the cognitive model it uses for this) defined in formal terms of the corresponding operating environment. Along with this, both the initial conditions for solving the problem (including fixed initial and final states) and the steps for solving it (business processes, executable scripts and scripts, and decision making rules) should also be expressed in terms of the same subject ontology.

Earlier study [6] shows the fundamental possibility of solving the same problem of experiential learning agnostic to the representation of the environment. That is, the same real problem can be successfully solved by various learning algorithms, based on agent's own acting experience, having the problem and its subject domain represented by different alternative ontological schemas. In this work, we assume the possibility of solving a wide range of problems using a cognitive architecture defined for an arbitrary operating environment, established by a corresponding ontological subject domain model. With regard to the classes of tasks associated with both business process management and psychotherapeutic practice [9], we propose the use of a basic process-oriented (or "activity") ontology [10] as a basis for describing specific applied ontologies. The "activity ontology" proposed in the mentioned work [10] suggests working with so-called "intelligent objects" and "intelligent functions" in terms of earlier work [2]. The main classes of intelligent objects are considered "invariants" and "instances" ("precedents") - the former are stable, expressed in space and time, representations of the latter, while the latter are the "evidence bases" for identifying the former. In turn, "intelligent functions" are represented by several classes of cognitive operations performed on "intelligent objects", as described further.

The implementation of the task-driven (TDA) approach described above on the basis of the previously listed five principles is supposed to be based on three computational methods – logical-probabilistic inference (LPI) [7], probabilistic formal concepts (PFC) [8] and the theory of functional systems (TFS) [3,4,5] as presented below.

*1.3. Logical-probabilistic inference*

The method of logical-probabilistic inference (LPI) [7,11], along with the Non-axiomatic Reasoning System (NARS) [12] and Probabilistic Logic Networks (PLN) [13], allows for solving cognitive problems with learning, recognition, and prediction. These problems are traditionally associated with machine learning methods based on various regression models, including those based on artificial neural networks (ANNs). The supposed advantage of these methods is the fundamental explainability of their predictions, as well as the interpretability of the models themselves, obtained in the course of training, insofar as they are all based on an ontological, that is, semantic representation of information, differing only in the mathematical models used to calculate probabilities. An additional advantage of the LPI is that this method, unlike the other listed methods, along with the ANNs themselves, allows for the modeling of the function of human brain neurons [14], as well as providing compatibility with other methods (PFC and TFS), which are described further.

The LPI method, similarly to NARS and PLN, is based on basic operations for calculating probabilities such as "revision" (accounting for accumulated experience for the formation of rules linking causes and effects), "deduction" (predictions based on gained experience, as well as deriving new rules based on existing experience), "induction" (predictions under uncertainty and the generation of hypotheses about possible consequences based on existing causes), and "abduction" (the generation of hypotheses about possible causes due to existing effects).

For example, as shown in Fig. 2, the accumulation of information allows for the association of malnutrition with a weakened immune system, which provides a basis for assessing the probability of a rule connecting the two conditions due to the "revision" operation. Further, the presence of a rule that a weakened immune system leads to an increased risk of infection makes it possible to predict an increase in the risk of infection in the case of poor nutrition by applying "deduction" to both of the above rules, and to create a new rule. In addition, the established association of poor nutrition with an increase in the risk of hypothermia allows us to construct a hypothesis about the relationship of hypothermia to the weakening of the immune system through the "induction" operation. Finally, an infection due to hypothermia, in the presence of information about a previous decrease in immunity, allows us to pose a hypothesis about the connection of the latter with hypothermia through "abduction".

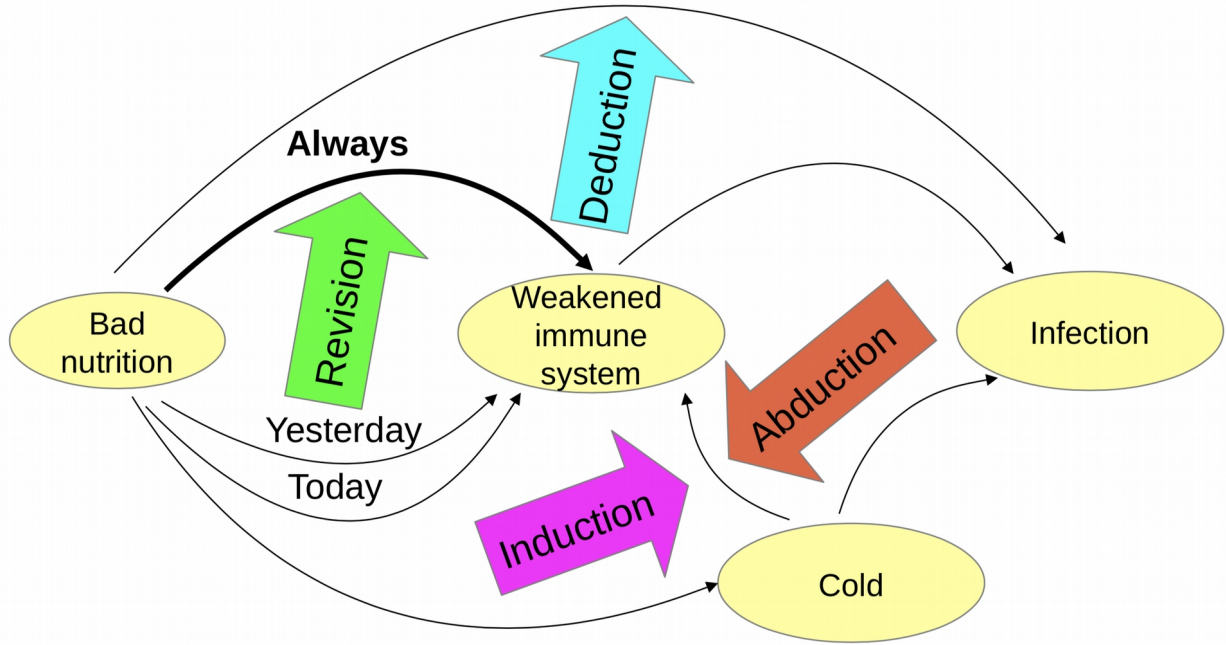

Fig. 2. An example of the use of various logical operations on an example from the subject area of healthcare.

*1.4. Probabilistic formal concepts*

The method of probabilistic formal concepts (PFC) [8,15] is an extension of the well-known method of formal concept analysis (Formal Concept Analysis) based on the LPI methodology described above. This method makes it possible to identify significant invariant combinations of features and contexts as interrelated causal associations, which can be interpreted as "fixed points" or "attractors" in terms of the neural network dynamics of Hopfield networks, based on repeatedly observed or perceived instances (including time-bound precedents). These invariants can be considered as classes of objects or phenomena corresponding to given precedents in the space of significant features identified in the process of applying this method. In addition, the method also potentially allows for building hierarchies of such classes as systems of concepts, as well as hierarchies of the features themselves as categories that define these concepts, including the values defined in scopes of domains of these categories.

At its core, the PFC is designated to solve the problem of clustering, which is widespread in the field of machine learning. It is supposed to do that within the framework of a transparent interpretable paradigm that allows both the verification of the semantic structure of the identified hierarchies as well as the association of each identified invariant with the rules that define it, as well as the verification of these rules themselves.

For example, in the context of the fruit recognition problem in Fig. 1, manually sorting and examining all the fruits on the counter with the processing of perception data by the PFC method will reveal high-level invariants such as fruits and berries, where fruits can include invariants of oranges, apples and pears, while apples can be split into invariants of green, yellow and red apples. At the same time, the significant features in this system of concepts will be weight, size, hardness and smoothness, as well as color, while the areas of definition of possible color values will be different for different top-level invariants (for example, citrus fruits can only be green, red, orange or yellow, but never blue).

*1.5. Theory of functional systems*

The applied development of the method based on the theory of functional systems (TFS) [3,4,5] is aimed at solving the generalization of the so-called "reinforcement learning" (RL) in machine learning. Traditionally, RL involves learning based on positive or negative feedback from the environment in the form of positive or negative reinforcement received upon solving certain problems, or failing to solve them. Most of the currently popular RL models are not explainable and, more importantly, not interpretable. At the same time, TFS allows extending the class of problems to cases where external reinforcement rarely appears explicitly, is delayed or does not exist at all and self-supervised learning is necessary. In TFS, this is achieved through the basic mechanism of reinforcing the system for achieving the expected result of the task, created at the beginning of the task and confirmed when it is evaluated by the acceptor of the task. This corresponds to "experiential learning" (EL) [6] which includes both self-learning by trial and error, and learning based on explicit or implicit positive or negative reinforcements from the environment and/or the teacher.

For example, in the case of self-learning to play ping-pong in the well-known formulation of the RL problem interpreted in terms of EL [6], the first stage of learning might include the formation of the ability to distinguish moving objects (ball and racket) from stationary ones (by the PFC method), the second stage is to learn how to control the racket, linking the use of motor functions with the re-arrangements of moving objects, the third stage is learning to hit the ball, linking the change in its trajectory with the spatial configuration of the racket relative to the ball, the fourth, and final, stage is learning the actual rules of the game necessary to obtain the final reinforcement upon winning.

## 2. Unified activity ontology

In order to ensure the possibility of building a line of applications based on the same cognitive architecture using the methods described above, we assert the presence of a simple upper (foundation) ontology [10] focused on practical activity, which allows us to describe metadata for the entire proposed line of applied applications, describing three of them as examples below.

*2.1. Upper (foundation) ontology model*

As a basis for the metadata structure, following OWL (https://www.w3.org/TR/owl-ref/) and Schema.org (https://schema.org/) standards, we accept the basic concepts of Thing and Property. However, the further hierarchy of classes underneath them is altered for our concept, taking into account our understanding of an intelligent object (IO) [1] as a universal Thing that has a set of Properties. In this understanding, following the concept of an activity or process-oriented ontology [10], IO-s can be unique Instances or be abstract Invariants each of the latter a set of unique instances.

Instances can be specific Values (numbers, dates, texts, any physical or biological or social objects) or Precedents of individual Events fixed in time, uniting the moments of Coincidences, as well as time-stretched Processes interconnected in time – uniting Events of different times and their Coincidences. Values can be represented by complex composite Entities that are unique in the domain space (for example, a well-known physical apple or a specific Client of a particular bank), specific atomic Categories (for example, the color red or the male gender) used to identify and describe Entities, or – arbitrary relations uniquely fixed in the subject area, uniting various entities (for example, kinship relations between specific Clients). Events can be subdivided into States that are fixed at specified points in time (for example, the balance of the Account or the age of the Client) or Actions that change these states (such as a financial transaction or a pizza delivery for a birthday). Invariants, in turn, are also divided into abstract Properties that are invariant to their Values and timeless invariants of Events, Coincidences and Processes. We call Event invariants Appearances, Process invariants – Scenarios, and Coincidence invariants can be Scenes and Forks. Scenes, as conjunctive sets, unite the Appearances corresponding to the Events that coincide many times together. Forks, as disjunctive sets, include alternative Appearances corresponding to different Events that are possible at the same stage of the same Scenario, recorded in different Processes (for example, cash payment or card payment within the scenario of buying goods in a store). At a lower level, the invariants of the Category values are the corresponding Classifiers (for example, the Color Property Classifier defines a domain with the domain values Red, Yellow, and Green), and the invariants of specific Entities are their abstract Images.

*2.2. Cognitive behavioral therapy model*

In a particular case of solving the problem of diagnosis within the framework of cognitive behavioral therapy (CBT) in the context of [9], from the point of view of the TDA, we consider the task of making a Diagnosis, which is an invariant that combines the values of such a number of Classifiers as Feeling, Emotion, State, Psycho-type, Social Situation and Cognitive Distortion [16], with the corresponding domain values – Categories, defined according to a particular psychotherapeutic school or practice. The task is solved within the framework of a psychotherapeutic session, recorded by the Session Precedent (Process subclass), while many Sessions may be conducted according to one Protocol (Scenario subclass). The solution of the task is carried out in a course of a dialogue between the Client and the Psychotherapist (subclasses of Entity – Instance) by performing Actions in the context of the Session. The Therapist can perform Diagnostic or Treatment Actions, and the Client can respond to them through Reactions. All of these actions can be characterized by Speech Patterns (eg., Cognitive Distortion patterns [16]) associated with certain Classifiers listed above. In this model, in the course of session interaction, the psychotherapist first performs Diagnostic Actions, stimulating the Client to Reactions, which make it possible to identify certain Categories, the totality of which, according to all used Classifiers, makes it possible to identify the Diagnosis invariant in case of session success, or not to identify any, in case of failure. It should be noted that the formal failure of the session in this formulation can mean several different things: a) the failure of the Psychotherapist to identify a real problem with the Client; b) the Client does not have a real problem; c) the successful resolution of the Client's problem during the session Interaction itself, especially if some of the Activities were Healing. In the last two cases, the TDA allows not only making a Diagnosis (in the case of formal success), but also to maximize the healing effect, ensuring the removal of all possible diagnoses in case of any and then verifying the absence of each.

*2.3. Enterprise customer relationship management model*

In the particular case of the formation of Offers to the Client of the Enterprise as part of Customer Relationship Management (CRM), the task, from the point of view of the TDA, is the formation of such an Offer, to which the Client will, with a high probability, respond positively, which will be recorded as a Subscription to the offered Product. Both the Offer and the Subscription are Actions in the proposed ontology, represented, as in the case of CBT, in the context of some short-term Interaction (the Client's contact with the Enterprise or a call to the Client from the Enterprise). An interaction can be a part of an Episode of Client-Enterprise interactions (like using a certain line of services in a certain office for a long time), and an Episode can be part of a longer-term client History (including all Episodes during the time when the Client used the Enterprise's services). The individual Clients themselves, as in the case of CBT, can be defined in the multidimensional space of a large number of Classifiers, such as Gender, Generation, Income, Social Status, Marital Status and many others, and can have links to Enterprise Products through the Current Subscription relationship. At the same time the stable combinations of Categories – the values of Classifiers, coupled with Current Subscriptions, can form such Invariants as the Client Image (for example, a young married man using a mortgage).

Both the Enterprise Client and the Product are subclasses of the Instance Entity, and the Offered Products that have Current Subscriptions are characterized by a Product Type. Histories, Episodes and Interactions are subclasses of Processes that accumulate historical records about Offers made to the Client and Subscriptions made by the Client in the context of the values of their Current Subscriptions, which can be recorded as Client States. The aggregation of this information makes it possible to identify Typical Histories, Typical Episodes and Typical Actions as Scenarios – Invariants, characterizing certain Client Images.

Based on the information available above, the task in terms of the TDA can be specified as generating Offers for such specific Products and Product Types, in the context of any given Customer, in order to maximize the likelihood of a Subscription to the offered Product, confirmed by the "action result acceptor" [3] by evaluating the results of delivering the Offer to the Client.

*2.4. Project management model*

A special case of a project management application involves solving the task of assigning a worker for a particular Task of a Project, where Tasks and Projects are Processes – Precedents, with the more atomic Tasks performed in the context of Projects. Both the Project and the Task can have an external assessment of "success" received from outside the system and fixed in the system logs, while the Project can have a current aggregated "success" metric evaluated before its completion based on the totality of the "success" of all Tasks included in it at the moment. Projects can be characterized in the space of such Classifiers as Audience and Subject Area, while the Tasks included in the corresponding Projects are defined in the space of such Classifiers as Priority, Severity, Status and Budget. Tasks are also associated with different Employees, who can occupy certain Positions (such as Director, Head of Department, Lead Engineer, etc.), and the positions they hold allow Employees to act in certain Roles in the contexts of both Projects (Project Role) and Tasks (Task Role).

The task (from the point of view of the TDA) in this context is to determine the Task (within the framework of the discussed ontology) for an Employee (in the corresponding Position), so that assigning them the appropriate Task Role maximizes both the "success" of solving the task and makes the entire Project, including this task, "successful" generally. For this purpose, historical precedents for solving similar Tasks in similar Projects by the same Employee, as well as other Employees, may be taken into account, based on final successes of those Tasks and Projects.

**3. Cognitive architecture for decision making**

Cognitive architectures based on ontological or semantic models and probabilistic logic primarily include the Non-Axiomatic Reasoning System (NARS) [12] and OpenCog [17]. In these systems, there are algorithmic capabilities similar to LPI in the form of non-axiomatic logic [11] and Probabilistic Logic Networks (PLN) [13]. However, they do not fully provide the possibilities offered by us in the form of the TDA, PFC, and TFS, as was shown in our previous works [3,6]. Within the framework of the architecture that we propose, it is assumed that all specific applications will use a universal "cognitive kernel" (CK) that is executed on the basis of private ontologies of the relevant subject domains and which will provide top-level applications with the use of the capabilities of the LPI, PFC and TFS within the framework of the TDA based on the "cognitive database" (CDB).

The CDB itself, as shown in Fig. 3 is intended for storing data and metadata based on the top-level ontology described above and can be specialized for other subject areas, providing an appropriate programming interface (API) for the cognitive functions of the CK – LPI, PFC, TFS and TDA.

The approximate composition of the CK can be described as a set of services described below that provide API functions both to each other and to "external" applications. Thus, it is possible to implement the system within the framework of a so-called "micro-service" architecture based on the services of the PLI, PFC, TFS and TDA based on a single CDB.

*3.1. Cognitive database*

The Cognitive Database (CDB) includes the following three components shown in Fig. 3 on the right side.

The Meta-data Base (MDB) is designed to store ontological schemas of Instances and Invariants (by describing their Types and Attributes), as well as domain Values of Classifiers. The full or partial caching of the MDB in an in-memory graph database or a specialized caching system for quick access at the application layer level is used to: a) obtain storage schemes for arbitrary instances and invariants based on types and their attributes; b) obtaining text values of data objects (most importantly the domain Values of Classifiers), as well as Metadata – Types and Attributes for visual operations on the user interface side.

The Data Object Base (DOB) is designed to maintain, within the structure described in the MDB above, both information about instances and their relationships with each other as recorded by applications, and their derived invariants, providing such functionality as a) saving precedents in relational normalization projections of the corresponding structure initial data; b) retrieving the precedents in relational normalization projections corresponding to the structure of the tasks being solved; c) retrieving and storing the invariants in the relational normalization projections corresponding to the structure of the tasks being solved.

The Rule Base (RB)is used for storing correspondences between the conjunctive-disjunctive trees of pre-conditions and the probabilistic estimates of the onset of post-conditions inferred by LPI, in relation to both precedents and invariants discovered by PFC, while ensuring a) the preservation of the identified rules b)the retrieval of relevant rules based on the given criteria for pre-conditions and post-conditions.

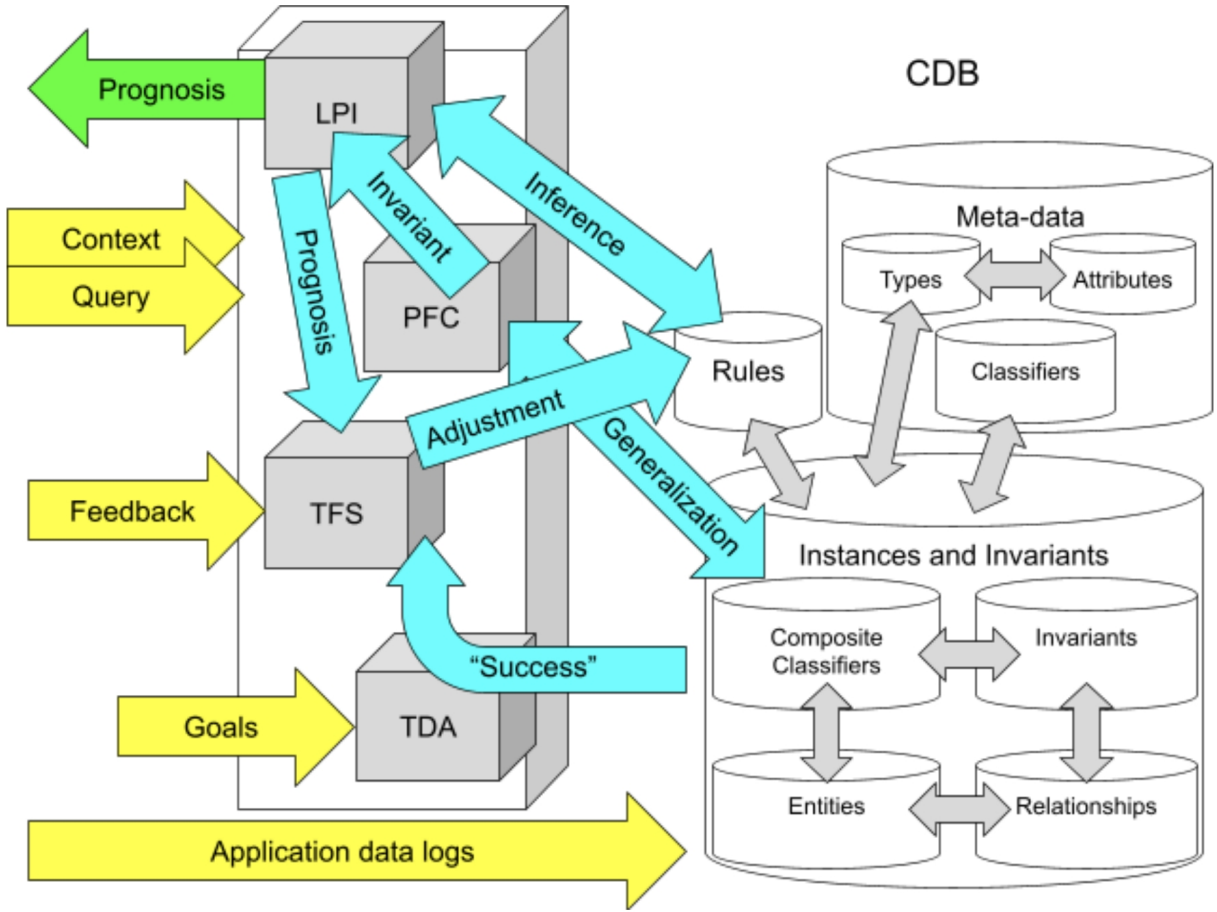

Fig. 3. Scheme of the cognitive kernel (CK), which includes the services of the LPI, PFC, TFS and TDA based on the CDB.

As follows from Figure 3, Types with Attributes in MDB describe the structure of Objects (Instances and Invariants) of the DOB, including the Relationships between them and Composite Classifiers, which can be created based on superpositions of the values of the original raw Classifiers, being derived on the basis of the identified Invariants by means of PFC, and the RB Rules can be associated with the Invariants for which they are revealed according to the LPI inference.

*3.2. Logical-probabilistic inference (LPI) service*

The LPI service implemented according to [7,11] provides for the following functionality: a) the formation of rules based on the Instances (and Precedents) recorded in the DOB and saving them in the RB; b) building forecasts and issuing recommendations based on the presented precedents, using both the rules found in the RB and the corresponding invariants found in the DOB c) using the PFC with the possibility of ranking the forecasts both by "probability" (in terms of [7,11]) and by "statistical significance" (Fischer's criterion), as well as by a certain combination of both. The provided forecasts can be transferred both to an external application system for practical use and to the TFS service to correct the model of cognitive behavior contained in the form of RB rules upon feedback.

The service can be controlled both by explicit (ad-hoc) requests from external applications and by background tasks designed to periodically update the RB. In this case, both explicit requests and background tasks can be accompanied by additional contextual information that refines the request.

*3.3. Probabilistic formal concepts (PFC) service*

The PFC service implemented according to [8,15] is intended for: a) identifying significant Invariants based on the data of Instances (and Precedents) in the DOB; b) preservation of the identified Invariants in the DOB.

The operation of the service can be managed as background tasks designed to periodically update the DOB Invariants and the associated BP Rules. At the same time, background tasks can be accompanied by additional contextual information specifying the search area for Invariants. Also, it is possible to use the service via direct (ad-hoc) requests from the LPI or application applications to identify Invariants in a certain required application context.

*3.4. Theory of functional system (TFS) service*

The TFS service implemented according to [2,3,4,5] provides the following functionality: a) receiving LPI forecasts and comparing them with actually recorded precedents in the DOB along with possible explicit feedback coming from the external environment (simultaneously logging the feedback in the DOB) in response to the issued LPI forecasts; b) adjusting the relevant Rules in the RB, depending on the results of comparing forecasts with their success. The action of TFS is in this sense similar to the "global feedback" [6] of currently active cognitive processes in the context of episodic memory in the case of perception of "successful" results.

The functioning of the service ensures the perception of both "external reactions" from the application system, based on the implementation of the recommendations of the LPV KY, and "internal incentives" generated by the TDA service.

*3.5. Task-driven approach (TDA) service*

The TDA service [3,4,5] is to be implemented as a specialized "Oracle", in relation to a specific subject area, which provides "external" (in relation to the rest of the architecture) "reinforcement" in accordance with the global "goals". Goals can be considered as given "from above" for a specific subject area through the definition of a "success function" calculated according to the posed goals and the criteria for their achievement based on data recorded by the entire system in the course of its operation. Evaluations of the "success" of "solving problems" ("achieving goals") are transferred to the TFS service as "internal" reinforcements that mediate external signals and the system's reactions to them in the context of the "success function".

The functioning of the service can be controlled both by triggers for certain events (such as the completion of Tasks or Projects), and by background tasks designed to periodically reinforce current activities through the TFS service.

## 4. Application examples

*4.1. Session diagnostics in cognitive behavioral therapy (CBT)*

Let's consider an example of an applied application – the "Psychotherapist" chatbot, designed to make a preliminary diagnosis of a Client who turned to a real psychotherapist for help, as shown in Fig.4. The application has an application part that provides the following: a) the actual functionality of the chatbot, connecting it with instant messengers such as Telegram and Slack; b) a module for detecting confirmations ("yes", "exactly", etc.), denials ("no", "by no means", etc.) and speech patterns (based on pattern detection technology [18] and/or the neural network implementation of a text classifier in the pattern data space based on semantic proximity [19,20]) to identify Speech Patterns in the Client's Reactions; c) the cognitive kernel (CN) itself based on the cognitive architecture described above to obtain the recommended diagnostic Speech Patterns for Diagnostic Influences made by the Psychotherapist in response to the Speech Patterns received from the Client, extracted from his Reactions; d) a module for generating textual Diagnostic Influences based on the patterns recommended by CK (implemented using interpretable text generation based on "connection grammar" [21] or pre-trained language models based on "transformers" [22]); e) the current Client context module, which tracks the history of the session, remembers all diagnostic actions (to prevent unnecessary repetitions) and stores the current diagnostic context, including the current excluded and confirmed classifier values.

P: Hi! How are you?
K: Bad. They have kicked out the loser! (tentatively diagnosing #being fired and #personalization)
P: How did that happen? Who did that? (clarifying **social situation**)
K: It was my job. I am fired so to speak. (confirming #being fired)
P: Does it make you anxious you or make you angry on them? (clarifying **emotion** - #anxiety or #anger)
K: Anxious, of course. I have to pay the mortgage. (confirming #anxiety)
P: You have been fired from your job and you are afraid, that won't be able to pay, right? (#being fired, #anxiety)
K: Yes. (confirming diagnosis in respect to **social situation** and **emotion**)
P: What do you think was the reason?
K: I just was not lucky – the investor has left the company. (diagnosing #personalization)
P: Was that your fault? (clarifying **cognitive distortion** - #personalization)
K: Not at all. I just feel like my life is ruined. (change diagnosis to #catastrophizing)
P: You are anxious, that after the being fired you won't be able to find a new job, correct? (#anxiety, #being fired, #catastrophizing)
K: Exactly. Although in fact I am an experienced professional. (confirming diagnosis in regard to **emotion**, **social situation** and **cognitive distortion** moving on to correction with healing interventions)

Fig. 4. An example of an imaginary protocol for automatic session diagnostics for CBT. P: - Psychotherapist replies, C: - client replies. Explanations of the actions of the designed system are given in parentheses. The hash tag # denotes the domain values (Categories) of such Classifiers as Emotion, Social Situation and Cognitive Distortion.

As can be seen from Fig. 4, the dialogue begins with a formal greeting of the Psychotherapist. The first reaction of the Client allows making a partial preliminary diagnosis in the space of two Classifiers - "personalization" [16] (Cognitive Distortion) upon "being fired" (Social Situation). The Psychotherapist confirms the hypothesis regarding the Social Situation and carries out an Influence to clarify the Client's Feelings. The Client's response allows the Feeling of "anxiety" to be identified, after which the Therapist confirms the partial diagnosis of "anxiety" in connection with "being fired" and proceeds to clarify the Cognitive Distortion. The Client's next reaction confirms the preliminary diagnosis ("personalization"), but an attempt to explicitly confirm this changes the diagnosis to "catastrophizing". At the end of the dialogue, the Psychotherapist confirms the full final diagnosis ("anxiety" in connection with the "catastrophe" of "being fired") and the Client confirms this, immediately giving a clue for correcting the situation through subsequent Healing Influences.

## 4.2. Formation of proposals in customer relationship management (CRM)

An example of the application of the proposed cognitive architecture in the area of the customer relationship management (CRM) system is the segmentation of the client audience. This can be done by means of predicting the qualities of the Clients, as described in [2], for the purpose of the formation of Product Offers for them on behalf of the Enterprise.

Fig. 5. Initial data for segmenting Clients and predicting their qualities in the space of the Classifiers of the customer relationship management system (CRM).

The appendix to [2] shows how the PFC-based statistical approximation algorithm can reveal expressive social contexts corresponding to the Invariants of individual Client Instances shown in Fig.5. Data on 2784 respondents (users) of the social network was studied in the context of 36 user characteristics, where each characteristic corresponds to a Classifier such as Gender, Place of Residence, Place of Birth, Values and Attitude to Bad Habits. As a result of the work of the statistical approximation algorithm based on the PFC, 21 Invariants were identified, which are distinguishable types of representatives of the user audience. The most salient of them were the 8th and 11th invariants. The 8th invariant can be characterized as such: married women living in their hometown, for whom the priority is family and children, who value kindness and honesty in people, with their relatives indicated in their profile, a negative attitude towards alcohol and smoking, a minimal number of photos, video and audio files. The 11th invariant, by contrast, can be characterized as such: unmarried men, perhaps teenagers, who also value kindness and honesty in people, however their main priority is self-improvement, with neutral stances on smoking and alcohol.

The segmentation of the Client audience into a number of Client Images, coupled with the recorded logs of the formation of Precedents of Offers of certain Products made to individual Clients in each of the segments, taking into account the record of subsequent Precedents of Client Subscriptions for the same products, may allow the LPI to make predictions about the acceptance or rejection of Offers for certain Products by any other Clients, taking into account the proximity of the Client to a particular Client Image and the probability of them accepting the Offer for each of the possible Products for this Client Image.

*4.3. Decision making in project management*

An example of the application of the suggested architecture is a personal assistant for a project manager that supports the decision-making processes based on LPI and TFS for a wide range of managerial tasks: the assignment of workers to the current Project Tasks according to Task Roles, for example. This can be provided by a hypothetical "Assistant Manager" service receiving the following inputs and generating the appropriate outputs, as described below.

The input consists of the operational contexts needed for issuing recommendations, which are automatically read by the drivers / plugins of existing corporate applications (messengers and task trackers) used by managers in their daily work and accumulated in the form of Precedents in the DOB as a continuous "event log" structured in the framework of the respective domain ontology that includes Employees, Positions, Task and Project Roles, as well as the Tasks and Projects themselves and all Classifiers related to this. In addition, the operational context can be manually entered by the manager themselves or by their human assistant.

The output presents recommendations of Employees for Positions to be assigned for a specific Task so that the manager independently makes the necessary decision selection from a list, ranked by "probability" and "statistical significance". The recommendations are inferred by the functions of the LPI and PFC which take into account the given context, in the form of a Task and its Status, as well as the lists of possible candidates from among the Employees and their Positions. It is also possible to automatically make decisions on the appointment of a worker – in this case, based on the ranked list of options provided by the LPI, an assignment can be automatically selected based on the top values of the "probability" and "statistical significance" values exceeding the "confidence threshold", set as either a global system setting or a user preference.

Decisions made in this way are recorded in the DOB as Precedents in the context of the current states of the Projects and Tasks and can later be evaluated within the TFS in terms of TDA regarding the success or failure of the tasks, reflecting on the results of the assessment for the purpose of adjusting the models used by the LIP for subsequent applications.

## 4. Conclusions and recommendations

We have shown how the five principles of brain activity (BPP) can be implemented within the framework of a task-driven approach (TDA) in a cognitive architecture built on the basis of methods of probabilistic inference (LPI), probabilistic formal concepts (PFC), and functional systems theory (TFS). We demonstrated the fundamental applicability of the proposed architecture for a number of applied tasks formalized in terms of the corresponding subject domain ontologies based on a basic activity ontology. In our further work, we plan to develop the proposed architecture and fully test it on real problems.

The paper is submitted for presentation at 2022 Annual International Conference on Brain-Inspired Cognitive Architectures for Artificial Intelligence.